\pdfoutput=1

\newif\ifarxiv
\arxivtrue

\newif\ifreviewmode
\reviewmodefalse %

\documentclass[11pt]{article}

\usepackage{acl}

\usepackage{times}
\usepackage{latexsym}

\usepackage[T1]{fontenc}

\usepackage[utf8]{inputenc}

\usepackage{microtype}

\usepackage{inconsolata}

\usepackage{graphicx}

\usepackage{amsmath,amsfonts,bm}

\usepackage{mathtools}
\usepackage{booktabs}
\usepackage{xspace}
\usepackage{tipa}

\usepackage{xcolor}

\usepackage{wrapfig}

\usepackage{caption}
\usepackage{subcaption}

\usepackage{algpseudocode}
\usepackage{algorithm}

\definecolor{caribbeangreen}{rgb}{0.0, 0.8, 0.6}
\definecolor{carrotorange}{rgb}{1.0, 0.615, 0.0}

\newcommand{\method}[0]{STEP\xspace}
\newcommand{\methodexplain}[0]{\textbf{S}yntactic \textbf{T}ransformation \textbf{E}nhanced \textbf{P}re-training\xspace}

\newcommand{\eg}{e.g.~}
\newcommand{\ie}{i.e.~}

\usepackage{multicol}
\usepackage{multirow}

\renewcommand{\paragraph}[1]{\textbf{#1}\xspace}

\usepackage{enumitem}
\setlist[itemize]{parsep=0pt,topsep=2pt,itemsep=0ex,partopsep=1ex,parsep=1ex, leftmargin=2ex}

\newcommand{\tightmath}{\abovedisplayskip=5pt
\belowdisplayskip=5pt}

\newcommand{\R}{\mathbb{R}}

\usepackage{forest}
\usepackage{tikz-dependency}
\usepackage{xcolor}

\newcommand{\ci}[1]{\scalebox{0.7}{\textcolor{gray}{\ensuremath{\pm #1}}}}

\usepackage{chngcntr}

\usepackage{hyperref}

\usepackage{url}
\usepackage[capitalise]{cleveref}

\renewcommand{\baselinestretch}{0.966}

\title{Strengthening Structural Inductive Biases by \\ Pre-training to Perform Syntactic Transformations}

\author{Matthias Lindemann$^1$ \and Alexander Koller$^2$ \and Ivan Titov$^{1,3}$ \\
$^1$ ILCC, University of Edinburgh,
$^2$ LST, Saarland University,
$^3$ ILLC, University of Amsterdam \\
{\small \texttt{m.m.lindemann@sms.ed.ac.uk}, \texttt{koller@coli.uni-saarland.de}, \texttt{ititov@inf.ed.ac.uk} }
}

\begin{document}
\maketitle

\begin{abstract}
Models need appropriate inductive biases to effectively learn from small amounts of data and generalize systematically outside of the training distribution. 
While Transformers are highly versatile and powerful, they can still benefit from enhanced structural inductive biases for seq2seq tasks, especially those involving syntactic transformations, such as converting active to passive voice or semantic parsing.
In this paper, we propose to strengthen the structural inductive bias of a Transformer by intermediate pre-training to perform synthetically generated syntactic transformations of dependency trees given a description of the transformation. 
Our experiments confirm that this helps with few-shot learning of syntactic tasks such as chunking, and also improves structural generalization for semantic parsing.
Our analysis shows that the intermediate pre-training leads to attention heads that keep track of which syntactic transformation needs to be applied to which token, and that the model can leverage these attention heads on downstream tasks.%
\ifarxiv %
\footnote{We release our code, data and model at \href{https://github.com/namednil/step}{https://github.com/namednil/step}.}
\else
\footnote{Code, data and model checkpoints will be made publicly available upon publication.}
\fi

\end{abstract}

\begin{figure*}[t]
    \centering
    \includegraphics[width=0.9\linewidth]{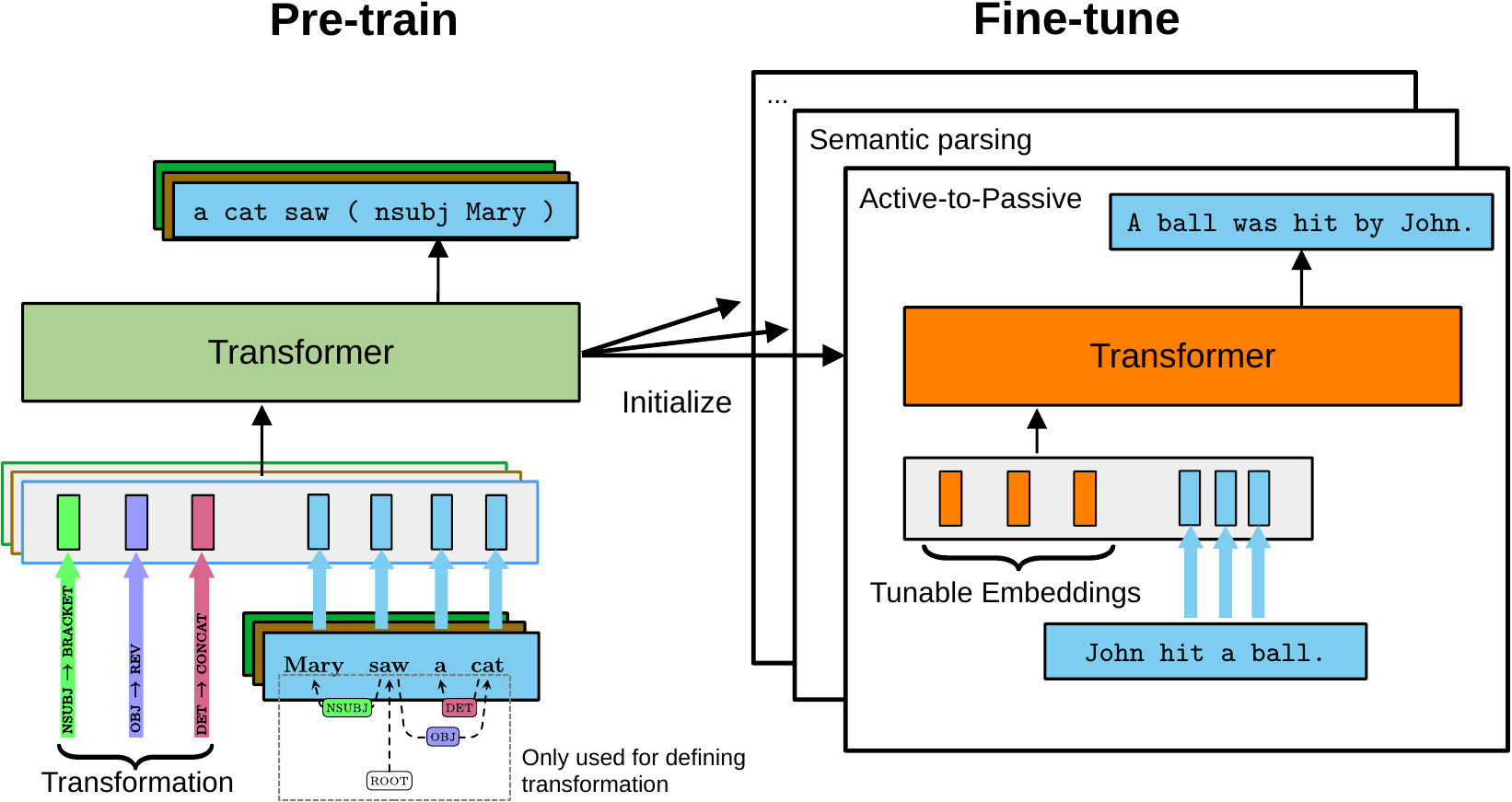}
    \caption{Left: Intermediate pre-training of a Transformer to perform syntactic transformations specified in the prefix; the syntax tree forms the basis of the transformation but is \textit{not} given to the model.
    Right: fine-tuning the Transformer and the prefix on a downstream task. Tunable parameters are represented in orange.}
    \label{fig:architecture}
    \vspace{-12pt}
\end{figure*}

\section{Introduction}

Inductive biases play a critical role in NLP, particularly in learning from limited data and in systematic generalization beyond the training distribution.
While standard seq2seq models excel on in-distribution data, they often lack structural inductive biases and hence perform poorly on structural generalization, \ie generalization to unseen combinations of known phrases \citep{keysers2020measuring}, extrapolation to longer inputs \citep{lake2018generalization, hupkes2020compositionality} and deeper recursion \citep{kim-linzen-2020-cogs, li-etal-2023-slog}.
While pre-training on large amounts of text improves structural generalization to a certain extent \citep{furrer2020compositional}, it remains challenging \citep{yao-koller-2022, li-etal-2023-slog}. 

This seems to conflict with observations that pre-training equips models with knowledge about syntax \citep{tenney-etal-2019-bert, hewitt-manning-2019-structural, mueller-etal-2022-coloring}, which should enable structural generalizations. 
In this paper, we start from the hypothesis that the lack of structural inductive bias is partly due to limited knowledge of how to \textit{use} syntactic information for structural tasks.

Traditionally, NLP has heavily relied on syntactic theories and has phrased many tasks as transformations of syntax trees, ranging from conversion of a sentence from active to passive voice \citep{oliva-1988-syntactic} to constructing a semantic representation for a sentence \citep{Montague1970-MONEAA-2}. Transformations of syntax trees can address a task in a very generalizable way by using the right abstractions.
For example, when constructing the semantic representation of an NP, by the principle of compositionality, the same transformations can be used for NPs whether they serve as direct objects or as indirect objects.

Inspired by this perspective, we propose a new method of strengthening the structural inductive bias of a pre-trained model with an additional intermediate pre-training step to perform syntactic transformations (see \cref{fig:architecture}). 
We create a dataset of automatically generated syntactic transformations of English dependency trees. Given a description of the transformation as a prefix and an input sentence, the model is pre-trained to predict the output of the transformation without access to the underlying dependency tree. 
This pre-training procedure encourages the model to strengthen its representations of syntax and acquire reusable dynamics of syntactic transformations that can be leveraged for downstream tasks.
During fine-tuning, gold-standard descriptions of transformations are not available, and we use a prefix of embeddings that are fine-tuned with the rest of the model instead.

\paragraph{Contributions} We demonstrate that our intermediate pre-training strengthens the structural inductive bias of the model, resulting in a better few-shot performance for syntax-dependent seq2seq tasks, such as conversion from active to passive or chunking. Our method also improves structural generalization in the context of semantic parsing.

Analysis of the pre-trained model shows that it uses attention heads to track what transformation needs to be applied to which input token, and that these heads tend to follow syntactic patterns. In addition, we find that fine-tuning re-uses these attention heads, suggesting that the model can leverage the transformations acquired during pre-training.

\section{Related Work}
\label{sec:relwork}

\paragraph{Pre-training with synthetic data}
Training on synthetic data to shape the inductive bias of Transformers has been explored in several recent works. \citet{papadimitriou2023pretrain} pre-train on a synthetic language to investigate the impact on language modelling of English. \citet{mccoy2023modeling} pre-train on a distribution of tasks using meta-learning \citep{finn2017model} and show improvements for low-resource language modelling of child-directed language.

Our work builds conceptually on SIP \citep{lindemann2023injecting}, in which a Transformer is pre-trained to simulate the behaviour of Finite State Transducers (FSTs) to introduce a structural inductive bias for FST-like behaviour. That is, given a representation of an automatically generated FST and an input string, a Transformer is pre-trained to predict what the output of the FST is on the given input. During fine-tuning, SIP proposes to use a prefix of tunable embeddings in place of an FST description.
While SIP and our work share similar methodology, \ie pre-training a model with a description of a transformation and fine-tuning the model with a prefix of tunable embeddings, they address different problems: SIP focuses on the sequential inductive bias of FSTs, whereas we strengthen the inductive bias for transformations of syntax trees. Another major difference is that their pre-training task is fully deterministic and unambiguous as there is only a single output for any FST and input string. In contrast, in our case, performing the transformation requires knowledge about the underlying syntax tree, which is not provided to the model. This forces the model to learn the syntax or reuse its existing syntactic knowledge.

\paragraph{Syntax-infused pre-training}
In recent years, several works have explored injecting syntactic knowledge through pre-training or multi-task learning.
Most of these approaches have focused on learning contextualized word representations with task-specific layers on top and have shown that syntactic knowledge can improve parsing \citep{zhou-etal-2020-limit}, semantic role labelling \citep{swayamdipta-etal-2018-syntactic, zhou-etal-2020-limit}, coreference resolution \citep{swayamdipta-etal-2018-syntactic}, grammatical error detection \citep{zhang-etal-2022-syntax} and relation extraction \citep{bassignana-etal-2023-silver}. Because these works focus on encoder-only models, they cannot be directly applied to sequence-to-sequence tasks.

In the context of sequence-to-sequence models, \citet{xu-etal-2020-improving} focus on broad-coverage semantic parsing and explore pre-training on multiple tasks including constituency parsing with linearized trees.
Finally, \citet{mulligan-etal-2021-structure} present proof-of-concept experiments in which they show that multi-task learning of syntactic transformations can provide a bias towards hierarchical generalizations when data with a hierarchical structure is provided for the auxiliary tasks. In contrast to our work, they consider a setup with training from scratch using multi-task learning rather than pre-training. They only use three manually selected syntactic transformations and focus entirely on synthetic data.

To our knowledge, we are the first to explore pre-training with a large space of synthetic transformations of syntax trees. In addition, rather than using an atomic and unstructured task id to distinguish different tasks \citep{johnson-etal-2017-googles, xu-etal-2020-improving, mulligan-etal-2021-structure}, we provide the model with an explicit description of the transformation.

\paragraph{Structural generalization} Several different approaches have been taken in recent works to improve the structural generalization of neural network models.
\citet{liu-etal-2021-learning-algebraic, kim-2021-nqscfg, weissenhorn-etal-2022-compositional, lindemann-etal-2023-compositional-generalization} and \citet{petit-etal-2023-structural} have proposed different specialized architectures that have structural inductive biases by design. While very effective, these approaches tend to be difficult to train if the `correct' task-specific syntactic analyses or alignments are not available, necessitating often complex and computationally expensive training algorithms. Since these approaches are also typically tailored to one or a few related tasks, architectures have to be redesigned when a new kind of task is considered. 

Some other works have explored data augmentation \citep{andreas-2020-good, qiu-etal-2022-improving, yang-etal-2022-subs} to improve structural generalization. Because data augmentation is task-specific, it needs to be repeated and potentially also adapted to every new task. Data augmentation inherently risks introducing errors and noise to the training data. In contrast, our approach pre-trains a model once to perform syntactic transformations and can then be fine-tuned for different downstream tasks.

\section{Strengthening Structural Inductive Bias}

\begin{figure*}
    \centering
    \includegraphics[width=0.85\linewidth]{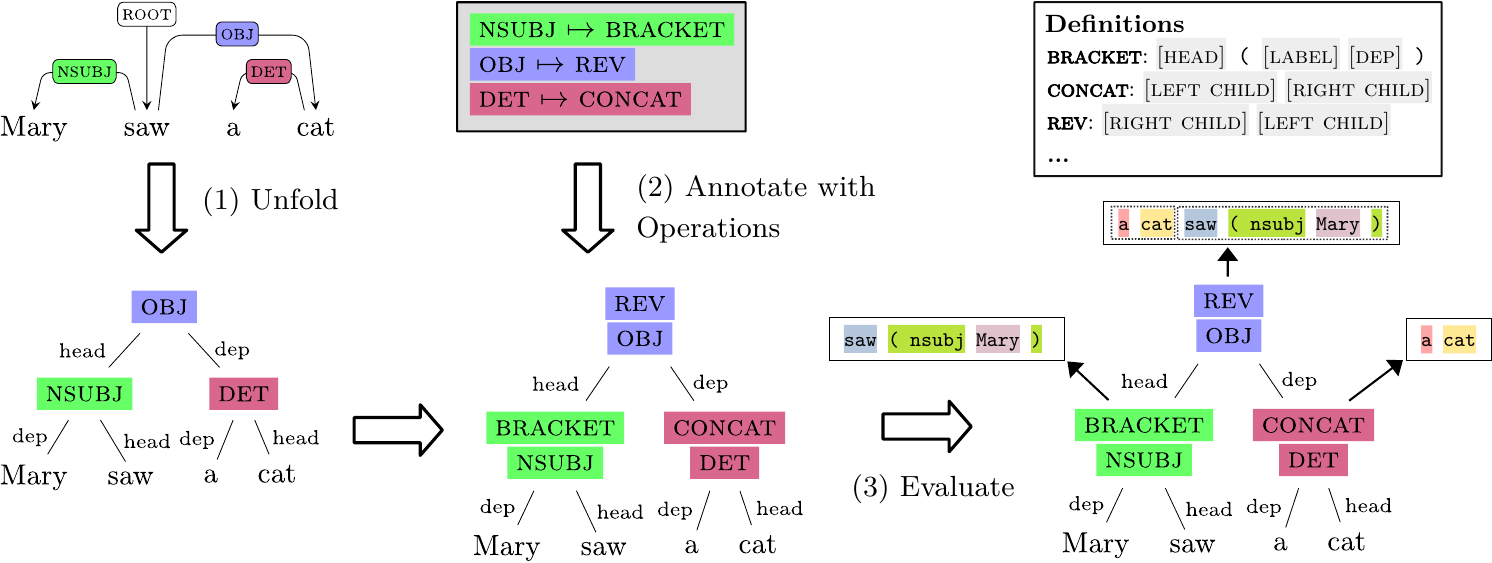}
    \vspace*{-2mm}
    \caption{Our procedure of applying a syntactic transformation specified as edgewise transformations (grey box): (1) recursively unfolding a dependency tree into a binary tree where dependency labels serve as labels of internal nodes, (2) annotation dependency relations with edgewise transformations, (3), recursive evaluation of the edgewise transformations with partial results shown.}
    \label{fig:transformation}
    \vspace{-6pt}
\end{figure*}

\begin{table*}[t]
    \centering
    \resizebox{\linewidth}{!}{
    \begin{tabular}{ll|l}
    \toprule
         \textbf{Name} & \textbf{Definition} & \textbf{Example} \\
    \midrule
\textsc{concat} & \colorbox{blue!30}{\textsc{left child}} \colorbox{orange!30}{\textsc{right child}} & \colorbox{blue!30}{Mary saw} \colorbox{orange!30}{a cat}\\
\textsc{rev} & \colorbox{orange!30}{\textsc{right child}} \colorbox{blue!30}{\textsc{left child}} & \colorbox{orange!30}{a cat} \colorbox{blue!30}{Mary saw}\\
\textsc{bracket} & \colorbox{cyan!30}{\textsc{head}} \texttt{(} \colorbox{yellow!30}{\textsc{label}} \colorbox{lime!50}{\textsc{dep}} \texttt{)} & \colorbox{cyan!30}{Mary saw} ( \colorbox{yellow!30}{obj} \colorbox{lime!50}{a cat} )\\
\textsc{br-invert} & \colorbox{lime!50}{\textsc{dep}} \texttt{(} \colorbox{yellow!30}{\textsc{label}} \texttt{by} \colorbox{cyan!30}{\textsc{head}} \texttt{)} &  \colorbox{lime!50}{a cat} ( \colorbox{yellow!30}{obj} by \colorbox{cyan!30}{Mary saw} )\\
\textsc{bracket-2} & \texttt{(} \colorbox{cyan!30}{\textsc{head}} \colorbox{yellow!30}{\textsc{label}} \colorbox{lime!50}{\textsc{dep}} \texttt{)} & ( \colorbox{cyan!30}{Mary saw} \colorbox{yellow!30}{obj} \colorbox{lime!50}{a cat} ) \\
\textsc{triple} & \colorbox{cyan!30}{\textsc{head}} \texttt{(} \colorbox{green!40}{\textsc{head.lemma}} \colorbox{yellow!30}{\textsc{label}} \colorbox{purple!20}{\textsc{dep.lemma}} \texttt{)} \colorbox{lime!50}{\textsc{dep}} & \colorbox{cyan!30}{Mary saw} ( \colorbox{green!40}{see} \colorbox{yellow!30}{obj} \colorbox{purple!20}{cat} ) \colorbox{lime!50}{a cat} \\
\textsc{ignore-dep} & \colorbox{cyan!30}{\textsc{head}} & \colorbox{cyan!30}{Mary saw} \\
    \bottomrule
    \end{tabular}
    }
    \caption{General overview of the operations we use. We show an example transformation for the sentence \textit{Mary saw a cat} where \colorbox{cyan!30}{\textsc{head} = \texttt{Mary saw}} and \colorbox{lime!50}{\textsc{dep} = \texttt{a cat}}. \textsc{head.lemma} (\textsc{dep.lemma}) refers to the lemma of the head (dependent) that the node in question was unfolded from (in the example: saw \scalebox{0.75}{$\xrightarrow{\text{obj}}$} cat). See \cref{tab:all-operations} for a full list of operations, including variants of those shown here.}
    \label{tab:overview-operations}
\vspace{-12pt}
\end{table*}

Standard pre-training objectives, \eg with denoising objectives \citep{raffel2020exploring}, encourage models to acquire syntactic knowledge but provide little information about syntactic \textit{transformations}, which are central to many syntactic and semantic seq2seq tasks. 
Our research hypothesis is that intermediate pre-training to perform transformations of syntax trees encourages the model to (i) strengthen its representations of the syntactic categories to which transformations can be applied (\eg subjects, objects) and (ii) acquire \textit{reusable} dynamics of transformations that are useful for downstream applications. 
By providing an explicit description of the transformation as a prefix, different transformations the model has learned during pre-training can be `activated' by the right choice of prefix. For this reason, we fine-tune the model with a prefix of tunable embeddings to make it easy to leverage these transformations on downstream tasks similar to SIP \citep{lindemann2023injecting}.

In addition to learning about transformations of trees, we also want the model to incorporate knowledge about the syntax of the underlying language (\ie English, in this case). Hence, we do not provide syntax trees to the model during training, which also enables us to perform inference and fine-tuning without a parser.

\subsection{Syntactic Transformations}
\label{sec:syntactic-transformations}

Our goal in designing the transformations is to create a family of syntactic transformations which resemble a broad class of real downstream transformations. %
We base our syntactic transformations on Universal Dependency trees \citep{de-marneffe-etal-2021-universal}, and provide an overview in \cref{fig:transformation}. Each transformation is fully specified by a set of \textit{edgewise transformations} that assign a binary string operation (\eg \textsc{bracket}) to a dependency relation (\eg \textsc{nsubj}).

Applying a syntactic transformation to a dependency tree is a three-step process: First, we \textit{unfold} the dependency tree into a binary `phrase-structure'-like tree, where the dependency labels act as labels of the internal nodes.\footnote{Related conversions from dependency to phrase structure trees have been explored in \citet{xia-palmer-2001-converting}.} This is necessary because all our operations are binary and we need a binary tree along which we can evaluate the operations. Second, we annotate the dependency labels with the corresponding operations according to the edge-wise transformations. Finally, we recursively evaluate each operation in the resulting expression tree, yielding a single output string.

\begin{figure}[t]
    \centering
    \includegraphics[width=0.85\linewidth]{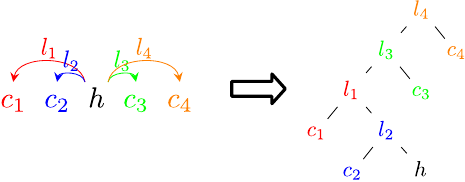}
    \caption{Unfolding a head $h$ and its children.}
    \label{fig:unfolding}
    \vspace{-12pt}
\end{figure}

Unfolding replaces a head and its dependents with a binarized tree, as shown in \cref{fig:unfolding}. This procedure is applied bottom-up to all nodes in the tree. For example, the dependency subtree of `a cat' unfolds to the tree $\textsc{det}(\textit{a}, \textit{cat})$, after which `saw' is unfolded, leading to the final unfolded result in \cref{fig:transformation}. Unfolding a node without children (\eg `Mary') simply retains that node.

In order to have a wide range of syntactic transformations, we design an inventory of 14 operations to cover many potentially useful transformations for downstream tasks (see \cref{tab:overview-operations} for a general overview, and \cref{tab:all-operations} for the full list).
Note that assigning the \textsc{concat} operation to all dependency relations results in an output that is identical to the input if the dependency tree is projective. 

\subsection{Intermediate pre-training}
\label{sec:pre-train}
During intermediate pre-training, the model is given a sentence and a set of edgewise transformations that determine the overall transformation. The objective is to predict what the transformation does to the parse tree of the sentence. The input to the Transformer is a sequence of vectors from $\R^d$, which consist of a prefix that represents the edgewise transformations and a suffix comprised of the embeddings of the input tokens:
\tightmath
\begin{align*}
\underbrace{\mathbf{h}_1, \mathbf{h}_2,  \ldots, \mathbf{h}_k}_{\text{Encoding of Transformation}}, \underbrace{\mathbf{x}_1, \mathbf{x}_2  \ldots, \mathbf{x}_n}_{\text{Sentence}}
\end{align*}
Each $\mathbf{h}_i$ encodes an edge-wise transformation $R \mapsto f$ by simple addition of embeddings:
\begin{align*}
    \mathbf{h}_i = \textsc{embed}_{\text{Label}}(R)+ \textsc{embed}_{\text{Transformation}}(f)
\end{align*}
The training objective is the log-likelihood of the correct output of the transformation, and we start from T5-base \citep{raffel2020exploring} that has already been pretrained. Note that the dependency tree is \textit{not} provided to the model to encourage it to reuse and strengthen its syntactic knowledge.

We also want to preserve the existing (syntactic) knowledge of the pre-trained T5 model, \eg to make it easy to insert the right auxiliary verb form when transforming a sentence from active to passive (see \cref{fig:architecture}). To help preserve this, our second pre-training objective is the span-denoising objective that T5 was originally pre-trained with. We train the model by alternating between gradient descent steps on the two objectives.

\paragraph{Data generation}
We construct random syntactic transformations for a small fraction of the C4 corpus, which T5 was originally pre-trained on. We tag, parse and lemmatize 2.1 million sentences with a total of around 39 million word forms using trankit \citep{nguyen-etal-2021-trankit}. We create two random transformations per parsed sentence, resulting in approximately 4.2 million pre-training instances.

To construct a random syntactic transformation for a given sentence, we sample dependency relations present in that sentence and some additional dependency relations that are \textit{not} present in the sentence to a maximum total of 20 relations. We uniformly sample an operation for each relation to create edgewise transformations.
Relations that are not chosen by sampling are implicitly assigned the operation \textsc{concat}. While the relations that are not present in the sentence have no bearing on the output of the transformation, we include them in the description to expose the model to a more general description that applies to a broader range of sentences.

\subsection{Fine-tuning}
\label{sec:fine-tune}
After pre-training, we apply our model to different downstream tasks via fine-tuning. Mirroring the pre-training, we replace the transformation encoding with a sequence of tunable embeddings. That is, the input to the model is a sequence of vectors:
\begin{align*}
\underbrace{\mathbf{h}_1', \mathbf{h}_2',  \ldots, \mathbf{h}_k'}_{\text{Tunable embeddings}}, \underbrace{\mathbf{x}_1, \mathbf{x}_2  \ldots, \mathbf{x}_n}_{\text{Sentence}}
\end{align*}
where $\mathbf{x}_1, \mathbf{x}_2  \ldots, \mathbf{x}_n$ are the embeddings of the input tokens, $\mathbf{h}_i' \in \mathbb{R}^d$ are the tunable embeddings and $k$ is a hyperparameter.
The embeddings $\mathbf{h}_i'$ are initialized to the average of the encoding of multiple transformations from the pre-training phase. Because the tuneable embeddings are trained on the downstream task, they can be used to `activate' transformations that help with the particular downstream task.
We fine-tune all model parameters and use a higher learning rate for the prefix.

\section{Evaluation}
We evaluate on syntactic and semantic tasks for which a structural inductive bias should be helpful. Specifically, we consider learning from a small amount of task-specific data (few-shot learning) and structural generalization outside of the training distribution to unseen combinations of known phrases, novel syntactic phenomena and deeper recursion than seen during training. 

\subsection{Baselines}
For a fair comparison, we compare our method (\method, for \methodexplain) with fine-tuning other seq2seq models based on T5-base \citep{raffel2020exploring} that were further pre-trained on the parsed corpus (\cref{sec:pre-train}) in different ways:

\paragraph{T5+Dep Parse} is pre-trained to predict a linearized dependency tree of the input, \eg \textit{Mary saw a cat} $\rightarrow$ \texttt{( saw nsubj Mary obj ( cat det a ) )}. Hence, this model incorporates syntactic information about English dependency trees but has limited exposure to how this information can be used other than to produce a parse tree.

\paragraph{Simple \method} is a simplified version of \method, where we always assign the same edgewise transformation to \textit{all} dependency relations. Consequently, the number of possible syntactic transformations is exactly the number of binary string operations we define (\cref{tab:all-operations}). However, we remove \textsc{ignore-dep} because it would result in an output string with a single token. We use a special token in the prefix to indicate which transformation should be applied. 

Analogously to \method, the models above were pre-trained with their specific pre-training objective and the original span denoising objective of T5.

\subsection{Syntactic Tasks}
\label{sec:eval-syn-transformation}
We first evaluate if our synthetic transformations transfer to realistic syntactic transformations.
In particular, we focus on few-shot scenarios.

We evaluate on three structural transformations that \citet{lyu-etal-2021-styleptb} identified as challenging because only several hundreds of training examples are available: passivization (\cref{fig:architecture}), emphasis of a designated adjective\footnote{\textit{The \underline{French} analysis goes further} $\rightarrow$ \textit{The analysis that goes further is French}} and emphasis of a designated verb\footnote{\textit{corporate profits may also \underline{dip} initially} $\rightarrow$ \textit{the dipping of corporate profits may also happen initially}}. We consider a more challenging version of this with only 100 training examples.

We report results in \cref{tab:vem-aem-atp} using exact match accuracy, BLEU \citep{papineni-etal-2002-bleu} and TER \citep{snover-etal-2006-study}, which is a normalized edit distance.
Using dependency parsing as the intermediate pre-training task (T5+Dep Parse) is already beneficial for passivization but somewhat deteriorates performance on adjective emphasis both in terms of BLEU and TER. Simple \method improves on this with small gains on both adjective emphasis and small additional improvements for passivization. \method performs best, outperforming the baselines by a sizable margin of 3.5 and 6 points BLEU on the adjective emphasis and passivization tasks. 
However, \method and T5 perform similarly on the verb emphasis task, and we hypothesize \method has difficulties reusing the transformations acquired during pre-training (see also \cref{sec:analysis}).

\begin{table}[t]
    \centering
\resizebox{0.97\linewidth}{!}{
\addtolength{\tabcolsep}{-0.4em}
\begin{tabular}{llrrr}
\toprule
 \textbf{Task} &  \textbf{Model} & \textbf{Acc} $\uparrow$ & \textbf{BLEU} $\uparrow$ & \textbf{TER} $\downarrow$ \\
\midrule
\multirow[c]{4}{*}{Verb emphasis} &  T5 & 3.5 & 41.7 & 46.7 \\
 & T5+Dep Parse & 3.3 & 40.1 & 48.2 \\
 & Simple \method & \textbf{3.6} & 40.5 & 47.0 \\
 & \method & 3.4 & \textbf{41.8} & \textbf{45.6} \\
\cline{1-5}
\multirow[c]{4}{*}{Adj. emphasis} & T5 & 7.3 & 47.6 & 38.3 \\
 & T5+Dep Parse & 7.7 & 45.8 & 40.4 \\
 & Simple \method & 9.8 & 48.3 & 37.5 \\
 & \method & \textbf{10.9} & \textbf{52.3} & \textbf{33.5} \\
\cline{1-5}
\multirow[c]{4}{*}{Passivization} & T5 & 40.2 & 73.7 & 18.3 \\
 & T5+Dep Parse & 45.0 & 76.8 & 15.5 \\
 & Simple \method & 46.8 & 78.4 & 13.6 \\
 & \method & \textbf{57.9} & \textbf{84.8} & \textbf{8.4} \\
\cline{1-5}
\end{tabular}
\addtolength{\tabcolsep}{0.4em}
}
    \caption{Evaluation on 100-shot \textbf{syntactic transformation} tasks. We report averages of 10 draws of 100 training examples each.}
    \label{tab:vem-aem-atp}
\vspace{-12pt}
\end{table}

\paragraph{Chunking} We also evaluate on chunking \citep{tjong-kim-sang-buchholz-2000-introduction} phrased as a seq2seq task.\footnote{\textit{The chairman promised Mr. Stone a decision $\rightarrow$   (NP The chairman) (VP promised) (NP Mr. Stone) (NP a decision)}}
Different variants of chunking play an important role in information extraction \citep{dong-etal-2023-open}, which often has to rely on small domain-specific corpora \citep{bassignana-plank-2022-crossre}. Few-shot learning of chunking is hence relevant and particularly interesting in our setup because it requires models to predict phrase categories (\eg NPs) that do not exist in our pre-training approach based on dependency trees.

We report results in \cref{tab:chunking}. While using parsing as intermediate pre-training is already helpful in comparison to T5, \method improves accuracy even further and outperforms T5 by almost 20 percentage points for exact match accuracy. Simple \method also shows some improvements over T5+Dep Parse but is again outperformed by \method.

Overall, this shows that \method strengthens the inductive bias for realistic syntactic transformations. The improvements of \method over T5 cannot be attributed alone to the prediction of dependency trees during pre-training as T5+Dep Parse performs worse. Pre-training the model with a narrow set of transformations (Simple \method) is not as effective as a large set of transformations with explicit descriptions. We hypothesize that the improvements of \method can be attributed partly to the reusability of the transformations during fine-tuning, which we analyze in \cref{sec:analysis}.

\begin{table}[t]
    \centering
\scalebox{0.95}{
\begin{tabular}{llrrr}
\toprule
\textbf{Model} & \textbf{Acc} $\uparrow$ & \textbf{F} $\uparrow$ \\
\midrule
T5 & $34.4\ci{0.8}$ & $87.4\ci{0.6}$ \\
T5+Dep Parse & $39.9\ci{2.1}$ & $90.0\ci{0.6}$ \\
Simple \method & $45.3\ci{2.0}$ & $90.6\ci{0.6}$ \\
\method & $\textbf{53.8}\ci{2.1}$ & $\textbf{93.2}\ci{0.5}$ \\
\bottomrule
\end{tabular}}
\caption{Means and standard deviations on \textbf{chunking} across 5 random draws of 100 training examples. Accuracy is exact match, \ie predicting \textit{all} chunks correctly.}
\label{tab:chunking}
\vspace{-12pt}
\end{table}

\subsection{Semantic Tasks}

\begin{table*}[t]
    \centering
\scalebox{0.92}{
\begin{tabular}{lrrrr|r}
\toprule
\textbf{Model} & \textbf{Modifiers} & \textbf{Novel Gaps} & \textbf{Wh-Questions} & \textbf{Recursion} & \textbf{Overall} \\
\midrule
AM-Parser$^*$ & $69.6$ & $20.7$ & $57.0$ & $\textbf{99.9}$ & $70.8\ci{4.3}$\\[0.5ex]
T5 & $79.5\ci{2.5}$ & $81.4\ci{6.2}$ & $82.4\ci{2.3}$ & $71.1\ci{1.2}$ & $77.6\ci{1.4}$\\
T5+Dep Parse & $82.8\ci{3.1}$ & $\textbf{86.8}\ci{6.6}$ & $78.8\ci{4.4}$ & $72.2\ci{2.0}$ & $78.3\ci{2.1}$\\
Simple \method & $\textbf{88.6}\ci{0.4}$ & $44.8\ci{12.6}$ & $84.2\ci{2.1}$ & $79.0\ci{1.6}$ & $78.8\ci{1.9}$\\
\method & $87.5\ci{0.4}$ & $47.6\ci{14.1}$ & $\textbf{84.8}\ci{4.2}$ & $79.8\ci{0.9}$ &$\textbf{79.3}\ci{2.3}$\\
\bottomrule
\end{tabular}}
\vspace*{-2mm}
\caption{Structural generalization on the variable-free meaning representation of \textbf{SLOG} based on 10 random seeds. $^*$ The AM-Parser uses a semantically more expressive meaning representation formalism based on graphs.}
\label{tab:slog}
\vspace{-12pt}
\end{table*}

\begin{table}[t]
    \centering
\scalebox{0.92}{
\begin{tabular}{llrrr}
\toprule
\textbf{Model} & \textbf{iid} & \textbf{Length} \\
\midrule
Tag \& Permute & $76.6\ci{1.7}$ & $\textbf{41.4}\ci{13.5}$ \\[0.5ex]
T5 & $\textbf{85.5}\ci{1.4}$ & $32.0\ci{4.4}$ \\
T5+Dep Parse & $84.9\ci{0.4}$ & $27.5\ci{2.0}$ \\
Simple \method & $82.9\ci{1.0}$ & $30.4\ci{1.5}$ \\
\method & $84.2\ci{1.7}$ & $38.4\ci{2.7}$ \\
\bottomrule
\end{tabular}}
\vspace*{-2mm}
\caption{Means and standard deviations of model accuracy for semantic parsing on \textbf{ATIS} for 5 random seeds.}
\vspace{-12pt}
\label{tab:atis}
\end{table}

Semantic parsing, \ie constructing a logical form from a sentence, can be seen as a particular transformation of the syntactic structure \citep{Montague1970-MONEAA-2}. Hence, we expect an inductive bias for syntactic transformations to be helpful for semantic parsing, particularly for \textit{structural generalization}, \ie extrapolation to unseen combinations of phrases, longer examples and deeper recursion. 

\paragraph{SLOG} \citep{li-etal-2023-slog} is a synthetic benchmark that tests models on 17 different structural generalizations grouped into 4 categories: using modifiers in novel positions (\eg PPs only modify objects during training but modify subjects at test time), novel gap positions (\eg wh-question for an indirect object, with the training data covering wh-questions for subjects and objects), wh-questions in novel syntactic contexts (\eg wh-questions combined with passive instead of active voice) and recursion (\eg deeper PP recursion).

We report aggregated results in \cref{tab:slog} and results for all 17 generalization cases in \cref{tab:slog-all}. Overall, \method performs best but performance on the different categories varies considerably between the approaches. \method and Simple \method outperform T5 on all but one category, with considerable margins for the novel modifier positions and unseen recursion depths. However, they underperform in the case of the novel gap positions. T5+Dep Parse performs more similarly to T5 with typically modest improvements across the categories.

The AM-Parser \citep{groschwitz-etal-2018-amr, weissenhorn-etal-2022-compositional} is a specialized approach for semantic parsing. It performs worse than the seq2seq models on most categories, except for recursion, where it achieves close to perfect accuracy.
Here, \method reduces the gap between the more general seq2seq models and the specialized AM-Parser. %
Interestingly, both \method and Simple \method improve over T5 on generalization to center embedding of depth 5 or more by 8 and 14 percentage points respectively even though there is no evidence of center embedding of depth two or more in our parsed corpus (\cref{tab:slog-all,fig:recursion-stats}).

\paragraph{ATIS} \citep{dahl-etal-1994-expanding} is a semantic parsing dataset with questions about a flight database annotated with executable logical forms. We follow previous work in using the variable-free FunQL version \citep{guo-etal-2020-benchmarking-meaning}. However, we found that the order of the conjuncts in the logical form tends to be somewhat unsystematic and often does not correspond to the linear order in the question. Hence, we use a pre-processing step to re-order conjuncts based on automatic alignments (see \cref{sec:preprocessing}).
We evaluate in two setups: (i) the standard iid split and (ii) a length split, where a model is shown logical forms with up to three conjuncts during training and has to generalize to sentences that require four or more conjuncts in the logical form.

Results are shown in \cref{tab:atis}. Tag \& Permute \citep{lindemann-etal-2023-compositional-generalization} is a specialized architecture for semantic parsing and is currently state-of-the-art on the length split. \method performs best among the non-specialized architectures on the length split, narrowing the gap to the specialized model. Interestingly, T5+Dep Parse and Simple \method perform somewhat worse than plain T5.

\section{Analysis}
\label{sec:analysis}

Our research hypothesis is that our intermediate pre-training encourages the model to acquire reusable dynamics of syntactic transformations that can be leveraged during fine-tuning.
In this section, we analyze the representations used by our model after its intermediate pre-training, and to what degree they are reused during fine-tuning.

\subsection{Analysis of Pre-Trained Model}
\label{sec:analysis:pre-train}
We first investigate how the model processes the transformation encoded in the prefix. The model has to attend to the prefix to gather information about which edgewise transformation needs to be applied to which input token. We call an attention head a \textit{transformation look-up head} if it consistently attends to the prefix.

We find that some transformation look-up heads are interpretable and follow syntactic patterns. For example, when head 6 in layer 10 computes the attention distribution for a token that is an object in the sentence (\ie \textit{cat} in \cref{fig:architecture}), it focuses the attention on the edgewise transformation that describes how to process objects (\ie $\textsc{obj} \mapsto \textsc{rev}$). 

\paragraph{Identifying interpretable look-up heads} We consider each attention head $H$ and dependency relation $R$ separately. For a sample of sentences with corresponding transformations, we count how many times the following conditions are true: (i) the instance has an edgewise transformation involving $R$, (ii) a token $x_i$ has an incoming edge labelled $R$ and (iii) $H$ focuses at least 50\% of its attention from $x_i$ on a single position $j$. If in over 70\% of cases, position $j$ refers to the edgewise transformation of $R$ then we call $H$ a transformation look-up head for the dependency relation $R$.

We find that there are often multiple transformation look-up heads per dependency relation. For example, we identify 7 look-up heads for \texttt{amod}. These interpretable heads are typically located in the mid or higher layers (see \cref{fig:dist-lookup-per-layer}), which is expected because the model first needs to identify the syntactic role each token has.

\paragraph{Intervening on look-up heads} Next, we verify that the transformation look-up heads we identified contribute to the model prediction with an interventional analysis.
We evaluate the role of the transformation look-up heads separately for different dependency relations: if the heads $H_1, H_2, \ldots H_n$ play an important role in performing transformations for dependency relation $R$, then masking all of them should drop accuracy for instances with an edgewise transformation for $R$. As a comparison, we also evaluate (i) masking $n$ randomly chosen heads, and (ii) masking $n$ randomly chosen heads that are transformation look-up heads, but not for $R$. 
Since the look-up heads can also have other functions within the model, we only mask out the attention to the prefix. When masking randomly selected attention heads, we ensure comparability by masking a random subset of tokens equal to the length of the prefix.

\begin{figure}[t]
    \centering
    \includegraphics[width=0.95\linewidth]{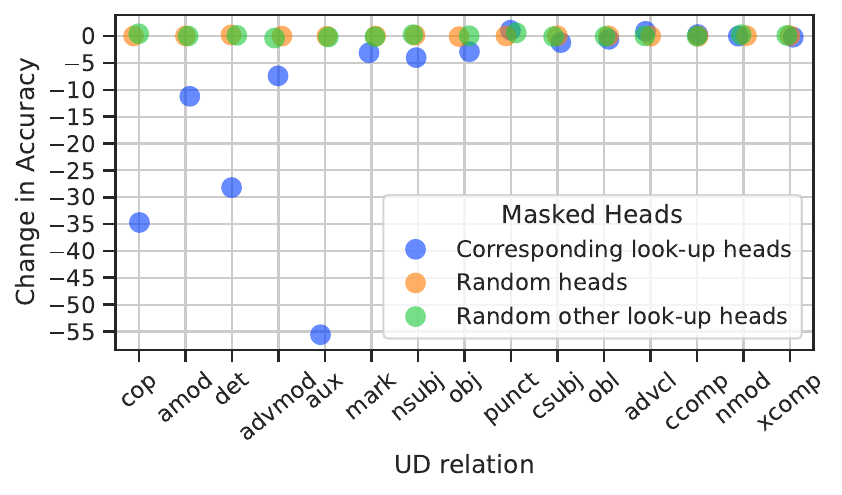}
    \vspace*{-3mm}
    \caption{Change in accuracy of predicting the output of edgewise transformations when masking different attention heads. We show accuracy relative to no masking.
    }
    \vspace{-12pt}
    \label{fig:mask-heads}
\end{figure}

We show the results in \cref{fig:mask-heads}. Masking transformation look-up heads reduces accuracy for many dependency relations while masking other transformation look-up heads or random heads has very little impact. This provides evidence that the identified heads play an important role within the model. For some relations (\eg punct, advcl, nmod), masking the respective look-up heads does not reduce accuracy, suggesting that responsibility for these relations is more spread out through the network.

\subsection{Analysis of Fine-Tuned Models}

How does a model pre-trained with \method learn during fine-tuning? We hypothesize that the pre-training provides a scaffolding, which finetuning can build upon. In particular, we expect that aspects of the downstream task that can be expressed with our transformations to be captured in the same way as during pre-training, \ie with the prefix and with the transformation look-up heads.

To test this hypothesis, we create 10 new synthetic transformation tasks with 5 edgewise transformations each and fine-tune the model (\cref{sec:fine-tune}). Then, we take the attention heads we identified in \cref{sec:analysis:pre-train} and repeat the masking intervention, \ie we mask the attention to the prefix of all look-up heads for the dependency relations involved in the edgewise transformations. 

\begin{figure}[t]
    \centering
    \includegraphics[width=0.9\linewidth]{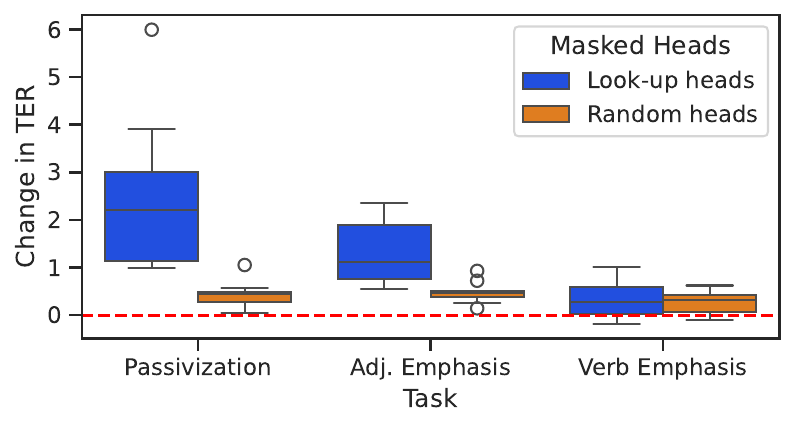}
    \vspace*{-3mm}
    \caption{Effect of masking look-up heads of models that have been fine-tuned on downstream syntactic tasks.
    For each task, we show the distribution for the 10 fine-tuned models from \cref{sec:eval-syn-transformation}.
    }
    \label{fig:masking-downstream-finetune-TER}
    \vspace{-12pt}
\end{figure}

Masking the look-up heads of the dependency relations involved in the transformations leads to an average drop in accuracy of 30 percentage points (see also \cref{fig:masking-simple-finetune}), whereas masking random look-up heads reduces accuracy by less than one point. We also find that one can \textit{read off} edgewise transformations from the learned prefix that agree with the ground-truth transformations with an average F-score of $\approx 77$ (see \cref{appendix:analysis}). 
This strongly supports the hypothesis that the model re-uses the transformations learned during pre-training and can `activate' them with the prefix.

\paragraph{Fine-tuning on realistic transformations} Finally, we investigate the role of transformation look-up heads in models fine-tuned on realistic syntactic transformations outside of the pre-training distribution (see \cref{sec:eval-syn-transformation}). Since there are no ground truth edgewise transformations in this case, we mask the attention to the prefix of \textit{all} transformation look-up heads and compare with masking an equal number of random heads. \cref{fig:masking-downstream-finetune-TER} shows that masking the transformation look-up heads deteriorates outputs more than masking random heads for passivization and the adjective emphasis task. However, results are comparable for verb emphasis. This is in line with our findings that \method improves over T5 for passivization and adjective emphasis but not for verb emphasis, and suggests that the lack of improvement for the verb emphasis task could be due to difficulties in reusing the transformations seen during intermediate pre-training.

\section{Conclusion}
We propose a new method of strengthening the structural inductive bias of a Transformer by pre-training the model to perform syntactic transformations based on dependency trees.
We show that this results in a better few-shot performance for syntax-dependent seq2seq tasks, and also improves structural generalization for semantic parsing.

Analysis of the pre-trained model shows that it uses attention heads to track what transformation needs to be applied to which input token, and that these heads tend to follow syntactic patterns. In addition, we find that fine-tuning re-uses these attention heads, suggesting that the model can leverage the transformations acquired during pre-training.

\section*{Limitations}

The structural inductive bias that is emphasized by our intermediate pre-training depends on the inventory of operations. Due to the computational cost of pre-training, we did not systematically explore which set of operations performs best, or which operations do not provide much benefit and could be omitted.

In this work, we focus on a moderately sized encoder-decoder model (T5) and do not investigate large decoder-only models. However, we do not foresee any reason why this approach could be less effective for such models.

Our analysis focuses on the encoder on the Transformer, and on the transformation look-up heads. However, applying a transformation also requires appropriate mechanisms in the decoder, and the picture of how this works internally remains much less clear.

\ifreviewmode
\else

\section*{Acknowledgements}
ML is supported by the UKRI Centre for Doctoral Training in Natural Language Processing, funded by the UKRI (grant EP/S022481/1), the University of Edinburgh, School of Informatics and School of Philosophy, Psychology \& Language Sciences, and a grant from Huawei Technologies. IT is supported by the Dutch National Science Foundation (NWO Vici VI.C.212.053).

\fi

\renewcommand{\baselinestretch}{1.0}

\bibliography{custom}

\appendix

\counterwithin{figure}{section}
\counterwithin{table}{section}

\section{Additional Details}

\subsection{Pre-processing}
\label{sec:preprocessing}
\paragraph{SLOG}
For SLOG, we remove \texttt{nmod.} from the logical forms to shorten them and to avoid giving models pre-trained with syntax trees a potential advantage simply because the downstream logical form uses a similar token to a dependency label. Hence, the logical form for `Isabella forwarded a box on a tree to Emma.' becomes \texttt{forward ( agent = Isabella , theme = box ( on = tree ) , recipient = Emma )} after pre-processing with the original one being \texttt{forward ( agent = Isabella , theme = box ( nmod . on = tree ) , recipient = Emma )}.

\paragraph{ATIS} We train an IBM-1 alignment model on the pairs of sentences and logical forms, and then sort the conjuncts by their sum-total expected alignment: let $A_{i,j}$ be the posterior probability that the input token at position $i$ is aligned to output the output at position $j$. Let $C$ be the set of output token positions belonging to a conjunct. We then let $A(C) = \sum_{j \in C}  \sum_i A_{i,j} \cdot i$. We repeat this for every conjunct and then sort them.

\subsection{Experimental setup}

\paragraph{Syntactic Transformations and Chunking} We evaluate in a few-shot scenario with only 100 training examples, and do not assume access to a development set. For this reason, we don't tune hyperparameters and fine-tune for a fixed number of epochs. 
As performance can differ from checkpoint to checkpoint, for each run, during the last 10 epochs, we evaluate on the test set and use the average result as the performance of that run.

For adjective emphasis, verb emphasis and passivization, we use all examples besides the 100 training examples as test set, \ie 2635 test examples for passivization, 596 for adjective emphasis, and 1101 for verb emphasis. For chunking, we use the test set from \citet{tjong-kim-sang-buchholz-2000-introduction}.

\paragraph{ATIS} We follow \citet{lindemann-etal-2023-compositional-generalization} in using the development set to select the best epoch based on the accuracy metric, which is also used on the test set (rather than loss).

\paragraph{SLOG} SLOG does not have an out-of-distribution development set, so we train for a fixed number of epochs that was determined by the hyperparameter search (see \cref{sec:hyper-params}).

\paragraph{Identifying look-up heads} We use a sample of 1000 unseen sentences from the C4 corpus along with randomly generated transformations as described in \cref{sec:syntactic-transformations} to identify interpretable look-up heads.

\paragraph{Intervening on look-up heads} Since we want to evaluate the impact of look-up heads for particular dependency relations, we create a dataset with 1000 examples of transformations per dependency relation. To avoid confounding factors, each instance has only a single edgewise transformation (for the specific dependency relation).

When we mask random attention heads or random look-up heads, it is computationally too expensive to do this for all possible attention heads and we approximate this with a Monte Carlo estimate: we select random heads 20 times and take the average of the results. 

\paragraph{Analysis of fine-tuned models on synthetic data} When generating synthetic downstream tasks, we exclude the \textsc{concat} operation for edgewise transformations. We take a sample of 5000 sentences from our parsed corpus and randomly divide it into an 80/20 train/test split.
We use a prefix of tunable embeddings of the same length as the ground truth, \ie we set it to a length of 5.
When masking random (look-up) heads, we repeat this 50 times to estimate the expected change in accuracy.

\subsection{Hyperparameters \& Hardware}
\label{sec:hyper-params}

\paragraph{Pre-training}
When generating pre-training data for \method, we only use sentences with 90 or less tokens (in terms of the T5 tokenizer) and exclude any instances with outputs of 180 T5 tokens or more. However, we do not impose a limit on the length of the output for our baselines (T5+Dep Parse and Simple \method) because it would remove too much of the pre-training.
We use Adafactor for our intermediate pre-training with a learning rate of $3 \times 10^{-4}$ and without warm-up, and a batch size of 80 (for \method), 30 (for Simple \method) and 48 (for T5+Dep Parse). We maintain separate optimizers for the main objective (\eg predicting the transformation) and the original span-denoising objective.
We train for a single epoch, except for T5+Dep Parse, which we train for two epochs. This is because \method and Simple \method have two instances with syntactic transformations per parsed sentence but T5+Dep Parse only has a single one.
For the denoising objective, we impose a limit of 80 tokens per instance (truncating longer instances) and use a batch size of 50.

\paragraph{Fine-tuning} During fine-tuning, the main hyperparameters are the learning rates. We use Adafactor for fine-tuning using a learning rate of $1 \times 10^{-4}$. For the prefix of \method, we use a learning rate of $10$. These hyperparameters apply to all experiments and all models, except for SLOG, as described below:

\paragraph{SLOG} We found that accuracy on SLOG was very sensitive to hyperparameters and used a hyperparameter selection strategy similar to that of \citet{conklin-etal-2021-meta} for COGS: we draw a sample of around 10\% from the generalization data. We fixed one random seed and ran 10 randomly sampled hyperparameter configurations and selected the one with the highest accuracy that was most stable across the epochs. We then discarded that random seed and used different ones for fine-tuning the model.
We sample the learning rate from $\text{LogUniform}[2 \times 10^{-6}, 1 \times 10^{-4}]$ and the batch size uniformly from $[24, 48, 72, 96, 120]$. \method also has an additional learning rate for the prefix, which we sample from $\text{LogUniform}[0.1, 10]$ during the search. The chosen hyperparameters can be found in \cref{tab:hparam-slog}.

\paragraph{Hardware} All our experiments were run on Nvidia 2080TI or 1080TI GPUs. Pre-training \method took around 30 hours. Since we used longer maximum sequence lengths for the baselines (see above), and had to decrease the physical batch size, training of the baselines took 50 (T5+Dep Parse) and 95 hours (Simple \method).

\paragraph{Number of parameters} T5-base has 222 million parameters. When we fine-tune \method with a prefix of tunable embeddings, this adds 7860 parameters to that, which is an increase of 0.035 \textperthousand.

\begin{table}[t]
    \centering
\resizebox{\linewidth}{!}{
\begin{tabular}{lrrrr}
\toprule
\textbf{Model} & \textbf{Epochs} & \textbf{Batch size} & \textbf{LR} & \textbf{LR Prefix} \\
\midrule
T5 & 50 & 96 & 1.62E-05 & - \\
T5+ Dep Parse & 50 & 24 & 9.51E-05 & - \\
Simple \method & 22 & 72 & 6.75E-05 & - \\
\method & 15 & 48 & 1.30E-05 & 2.52 \\
\bottomrule
\end{tabular}
}
    \caption{Hyperparameters used for \textbf{SLOG}. LR is learning rate.}
    \label{tab:hparam-slog}
\end{table}

\subsection{Evaluation metrics}

We use SacreBLEU \citep{post-2018-call} v2.3 to compute BLEU and TER. For the experiments with ATIS, we use the code of \citet{lindemann-etal-2023-compositional-generalization} for computing accuracy.

\paragraph{SLOG} \citet{li-etal-2023-slog} argue for using semantic equivalence for evaluation but they focus on a variable-based formalism and use exact string match for evaluating the variable-free representation. We take semantic equivalence into account, in particular, the order of the children does not matter because the roles are represented in the logical form. Hence, \texttt{offer ( theme = donut , recipient = * turtle )} and \texttt{offer ( recipient = * turtle , theme = donut )} are equivalent. We achieve this by parsing the string representation into a tree and instead of a list of children we maintain a set of children, and then compare trees to evaluate accuracy.

\begin{table*}[t]
    \centering
    \resizebox{\linewidth}{!}{
    \addtolength{\tabcolsep}{-0.35em}
    \begin{tabular}{l|l|l}
    \toprule
         \textbf{Name} & \textbf{Definition} & \textbf{Example} \\
    \midrule
\textsc{concat} & \colorbox{blue!30}{\textsc{left child}} \colorbox{orange!30}{\textsc{right child}} & \colorbox{blue!30}{Mary saw} \colorbox{orange!30}{a cat}\\
\textsc{rev} & \colorbox{orange!30}{\textsc{right child}} \colorbox{blue!30}{\textsc{left child}} & \colorbox{orange!30}{a cat} \colorbox{blue!30}{Mary saw}\\
\textsc{concat-rel} & \colorbox{blue!30}{\textsc{left child}} \colorbox{yellow!30}{\textsc{label}}  \colorbox{orange!30}{\textsc{right child}} & \colorbox{blue!30}{Mary saw} \colorbox{yellow!30}{obj}  \colorbox{orange!30}{a cat}\\
\textsc{revl-rel} &  \colorbox{orange!30}{\textsc{right child}} \colorbox{yellow!30}{\textsc{label}}  \colorbox{blue!30}{\textsc{left child}}  &  \colorbox{orange!30}{a cat} \colorbox{yellow!30}{obj}  \colorbox{blue!30}{Mary saw} \\
\textsc{bracket} & \colorbox{cyan!30}{\textsc{head}} \texttt{(} \colorbox{yellow!30}{\textsc{label}} \colorbox{lime!50}{\textsc{dep}} \texttt{)} & \colorbox{cyan!30}{Mary saw} ( \colorbox{yellow!30}{obj} \colorbox{lime!50}{a cat} )\\
\textsc{br-invert} & \colorbox{lime!50}{\textsc{dep}} \texttt{(} \colorbox{yellow!30}{\textsc{label}} \texttt{by} \colorbox{cyan!30}{\textsc{head}} \texttt{)} &  \colorbox{lime!50}{a cat} ( \colorbox{yellow!30}{obj} by \colorbox{cyan!30}{Mary saw} )\\
\textsc{bracket-2} & \texttt{(} \colorbox{cyan!30}{\textsc{head}} \colorbox{yellow!30}{\textsc{label}} \colorbox{lime!50}{\textsc{dep}} \texttt{)} & ( \colorbox{cyan!30}{Mary saw} \colorbox{yellow!30}{obj} \colorbox{lime!50}{a cat} ) \\
\textsc{bracket-2-inv} & \texttt{(} \colorbox{lime!50}{\textsc{dep}} \colorbox{yellow!30}{\textsc{label}} \colorbox{cyan!30}{\textsc{head}}  \texttt{)} & ( \colorbox{lime!50}{a cat} \colorbox{yellow!30}{obj} \colorbox{cyan!30}{Mary saw} ) \\
\textsc{bracket-3} & \colorbox{cyan!30}{\textsc{head}} \texttt{(} \colorbox{lime!50}{\textsc{dep}} \texttt{)} & \colorbox{cyan!30}{Mary saw} ( \colorbox{lime!50}{a cat} )\\
\textsc{bracket-4} & \colorbox{cyan!30}{\textsc{head}} \colorbox{yellow!30}{\textsc{label}} \texttt{(} \colorbox{lime!50}{\textsc{dep}} \texttt{)} & \colorbox{cyan!30}{Mary saw} \colorbox{yellow!30}{obj} ( \colorbox{lime!50}{a cat} )\\

\textsc{bracket-5} &
    \ensuremath{\begin{cases*}
    \colorbox{cyan!30}{\textsc{head}} \texttt{(} \colorbox{yellow!30}{\textsc{label}} \colorbox{lime!50}{\textsc{dep}} \texttt{)}  & \text{ if head has no other \textsc{bracket-5} arguments} \\
    \colorbox{cyan!30}{\textsc{head}} \texttt{(} \colorbox{yellow!30}{\textsc{label}} \colorbox{lime!50}{\textsc{dep}}  & \text{ if this is the first \textsc{bracket-5} argument} \\
    \colorbox{cyan!30}{\textsc{head}} \texttt{,} \colorbox{yellow!30}{\textsc{label}} \colorbox{lime!50}{\textsc{dep}} \texttt{)}  & \text{ if this is the last \textsc{bracket-5} argument} \\
    \colorbox{cyan!30}{\textsc{head}} \texttt{,} \colorbox{yellow!30}{\textsc{label}} \colorbox{lime!50}{\textsc{dep}} & \text{ else} \\
    \end{cases*}} & \colorbox{cyan!30}{Mary saw} ( \colorbox{yellow!30}{obj} \colorbox{lime!50}{a cat} )\\

\textsc{triple} & \colorbox{cyan!30}{\textsc{head}} \texttt{(} \colorbox{green!40}{\textsc{head.lemma}} \colorbox{yellow!30}{\textsc{label}} \colorbox{purple!20}{\textsc{dep.lemma}} \texttt{)} \colorbox{lime!50}{\textsc{dep}} & \colorbox{cyan!30}{Mary saw} ( \colorbox{green!40}{see} \colorbox{yellow!30}{obj} \colorbox{purple!20}{cat} ) \colorbox{lime!50}{a cat} \\
\textsc{triple-inv} & \colorbox{cyan!30}{\textsc{head}} \texttt{(} \colorbox{purple!20}{\textsc{dep.lemma}} 
 \colorbox{yellow!30}{\textsc{label}} \texttt{by} \colorbox{green!40}{\textsc{head.lemma}}  \texttt{)} \colorbox{lime!50}{\textsc{dep}} & \colorbox{cyan!30}{Mary saw} ( \colorbox{purple!20}{cat} \colorbox{yellow!30}{obj} \texttt{by} \colorbox{green!40}{see} ) \colorbox{lime!50}{a cat} \\
\textsc{ignore-dep} & \colorbox{cyan!30}{\textsc{head}} & \colorbox{cyan!30}{Mary saw} \\
    \bottomrule
    \end{tabular}
    }
    \addtolength{\tabcolsep}{0.35em}
    \caption{Full list of operations we use. We show an example transformation for the sentence \textit{Mary saw a cat} where \colorbox{cyan!30}{\textsc{head} = \texttt{Mary saw}} and \colorbox{lime!50}{\textsc{dep} = \texttt{a cat}}. \textsc{head.lemma} (\textsc{dep.lemma}) refers to the lemma of the head (dependent) that the edge in question was unfolded from (in the example: saw $\xrightarrow{\text{obj}}$ cat). 
    \textsc{bracket-5} essentially concatenates the results of all other \textsc{bracket-5} children together using a comma as joining element, and surrounds this with one matching pair of brackets. If in the example, we had edgewise transformations \textsc{nsubj} $\mapsto$ $\textsc{bracket-5}$ and \textsc{obj} $\mapsto$ $\textsc{bracket-5}$, the output would be \texttt{saw ( nsubj Mary , obj a cat )}, similar to our linearization of dependency trees for T5+Dep Parse.
    Formally, we call a subtree an $\ell$ argument in the unfolded and annotated tree if it is a non-head child that is dominated by a node that is annotated with operation $\ell$. For example, in \cref{fig:transformation}, the subtree corresponding to `a cat' is a \textsc{concat} argument.
    }
    \label{tab:all-operations}
\end{table*}

\section{Additional Results}

\begin{table*}[t]
    \centering
\begin{tabular}{lrrrr}
\toprule
\textbf{Generalization} & \textbf{\method} & \textbf{Simple \method} & \textbf{T5} & \textbf{T5+Dep Parse} \\
\midrule
Deeper CP tail recursion & $\textbf{74.5}\ci{4.2}$ & $73.0\ci{8.4}$ & $42.5\ci{4.4}$ & $50.9\ci{9.4}$ \\
Deeper PP recursion & $\textbf{87.3}\ci{2.5}$ & $78.5\ci{4.4}$ & $75.0\ci{4.5}$ & $70.8\ci{5.9}$ \\
Deeper center embedding & $17.3\ci{2.8}$ & $\textbf{22.7}\ci{2.2}$ & $8.9\ci{0.4}$ & $11.5\ci{2.1}$ \\
Shallower CP tail recursion & $\textbf{100.0}\ci{0.0}$ & $\textbf{100.0}\ci{0.0}$ & $\textbf{100.0}\ci{0.0}$ & $\textbf{100.0}\ci{0.0}$ \\
Shallower PP recursion & $\textbf{100.0}\ci{0.0}$ & $\textbf{100.0}\ci{0.0}$ & $99.9\ci{0.2}$ & $\textbf{100.0}\ci{0.0}$ \\
Shallower center embedding & $\textbf{100.0}\ci{0.0}$ & $\textbf{100.0}\ci{0.0}$ & $\textbf{100.0}\ci{0.0}$ & $\textbf{100.0}\ci{0.0}$ \\
\midrule
PP in indirect object NPs & $99.5\ci{0.4}$ & $99.8\ci{0.1}$ & $\textbf{100.0}\ci{0.0}$ & $99.7\ci{0.1}$ \\
PP in subject NPs & $\textbf{95.3}\ci{0.1}$ & $\textbf{95.3}\ci{0.1}$ & $74.5\ci{8.2}$ & $95.3\ci{0.2}$ \\
RC in indirect object NPs & $65.7\ci{0.8}$ & $70.7\ci{0.8}$ & $\textbf{73.0}\ci{0.8}$ & $67.1\ci{1.2}$ \\
RC in subject NPs & $\textbf{89.4}\ci{0.9}$ & $88.4\ci{1.4}$ & $70.4\ci{2.6}$ & $69.1\ci{12.3}$ \\
\midrule
Indirect object-extracted RC & $16.5\ci{13.0}$ & $11.7\ci{13.9}$ & $62.9\ci{12.3}$ & $\textbf{73.8}\ci{13.2}$ \\
Indirect object wh-questions & $78.8\ci{18.3}$ & $77.8\ci{14.6}$ & $\textbf{100.0}\ci{0.1}$ & $99.8\ci{0.2}$ \\
\midrule
Direct object wh-questions & $83.4\ci{14.5}$ & $91.0\ci{4.5}$ & $\textbf{99.2}\ci{0.6}$ & $83.8\ci{18.0}$ \\
Wh-questions long movement & $\textbf{55.8}\ci{12.5}$ & $46.2\ci{7.5}$ & $40.4\ci{8.1}$ & $35.2\ci{6.2}$ \\
Wh-questions with modified NPs & $\textbf{85.9}\ci{2.7}$ & $85.2\ci{1.9}$ & $72.9\ci{4.8}$ & $77.7\ci{2.8}$ \\
Active subject wh-questions & $\textbf{100.0}\ci{0.1}$ & $99.1\ci{1.8}$ & $99.6\ci{0.3}$ & $97.1\ci{5.6}$ \\
Passive subject wh-questions & $98.8\ci{2.4}$ & $99.5\ci{0.6}$ & $\textbf{100.0}\ci{0.0}$ & $\textbf{100.0}\ci{0.0}$ \\

\midrule
Average & $\textbf{79.3}\ci{2.3}$ & $78.8\ci{1.9}$ & $77.6\ci{1.4}$ & $78.3\ci{2.1}$ \\
\bottomrule
\end{tabular}
    \caption{Full SLOG results.}
    \label{tab:slog-all}
\end{table*}

\begin{figure}[t]
    \centering
    \includegraphics[width=\linewidth]{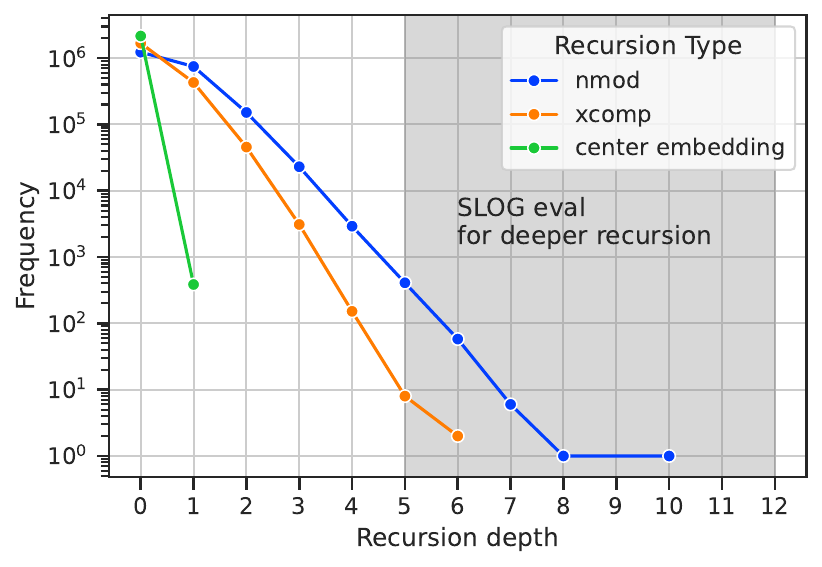}
    \caption{Frequency of recursion depths in our parsed corpus (\cref{sec:pre-train}) according to the dependency trees produced by trankit. Note that the y-axis is in log scale. In phrase structure terminology (\eg on SLOG), \textit{xcomp} recursion includes CP recursion and \textit{nmod} recursion includes PP recursion.}
    \label{fig:recursion-stats}
\end{figure}

\begin{table}[t]
    \centering
\resizebox{\linewidth}{!}{
\addtolength{\tabcolsep}{-0.4em}
\begin{tabular}{llrrr}
\toprule
 \textbf{Task} &  \textbf{Model} & \textbf{Acc} $\uparrow$ & \textbf{BLEU} $\uparrow$ & \textbf{TER} $\downarrow$ \\
\midrule
\multirow[c]{4}{*}{Adj. emphasis} & \method & $10.9\ci{1.0}$ & $52.3\ci{0.7}$ & $33.5\ci{0.6}$ \\
 & Simple \method & $9.8\ci{0.8}$ & $48.3\ci{0.8}$ & $37.5\ci{0.9}$ \\
 & T5 & $7.3\ci{0.8}$ & $47.6\ci{0.8}$ & $38.3\ci{0.8}$ \\
 & T5+Dep Parse & $7.7\ci{0.8}$ & $45.8\ci{0.9}$ & $40.4\ci{1.1}$ \\
\cline{1-5}
\multirow[c]{4}{*}{Passivization} & \method & $57.9\ci{2.0}$ & $84.8\ci{0.7}$ & $8.4\ci{0.5}$ \\
 & Simple \method & $46.8\ci{2.2}$ & $78.4\ci{0.8}$ & $13.6\ci{0.6}$ \\
 & T5 & $40.2\ci{1.7}$ & $73.7\ci{0.8}$ & $18.3\ci{0.7}$ \\
 & T5+Dep Parse & $45.0\ci{1.6}$ & $76.8\ci{0.5}$ & $15.5\ci{0.4}$ \\
\cline{1-5}
\multirow[c]{4}{*}{Verb emphasis} & \method & $3.4\ci{0.4}$ & $41.8\ci{0.6}$ & $45.6\ci{0.4}$ \\
 & Simple \method & $3.6\ci{0.7}$ & $40.5\ci{0.6}$ & $47.0\ci{0.6}$ \\
 & T5 & $3.5\ci{0.4}$ & $41.7\ci{0.5}$ & $46.7\ci{0.4}$ \\
 & T5+Dep Parse & $3.3\ci{0.7}$ & $40.1\ci{0.8}$ & $48.2\ci{0.6}$ \\
\bottomrule
\end{tabular}
\addtolength{\tabcolsep}{0.4em}
}
    \caption{Evaluation on 100-shot \textbf{syntactic transformation} tasks. We report averages of 10 draws of 100 training examples each. We also include standard deviations on the results across the 10 runs.}
    \label{tab:vem-aem-atp-full}
\vspace{-12pt}
\end{table}

\subsection{Analysis}
\label{appendix:analysis}

\paragraph{Decoding fine-tuned prefix} The importance of the look-up heads in the fine-tuned model suggests that the model uses the tunable prefix to encode task-specific information about which edgewise transformation to apply. To gain insight into this, we try to extract edgewise transformations from the fine-tuned prefix: 
For each vector $\mathbf{h}'$ in the prefix, we find the edgewise transformation whose embedding is closest to $\mathbf{h}'$ in terms of cosine similarity. In this manner, we can read off a candidate for the transformation which the model might be using under the hood and compare it to the correct transformation that generated the data. We find that the edgewise transformations extracted in this way agree with the gold edgewise transformations with an average F-score of $\approx 77$. 

\begin{figure}[t]
    \centering
    \includegraphics[width=\linewidth]{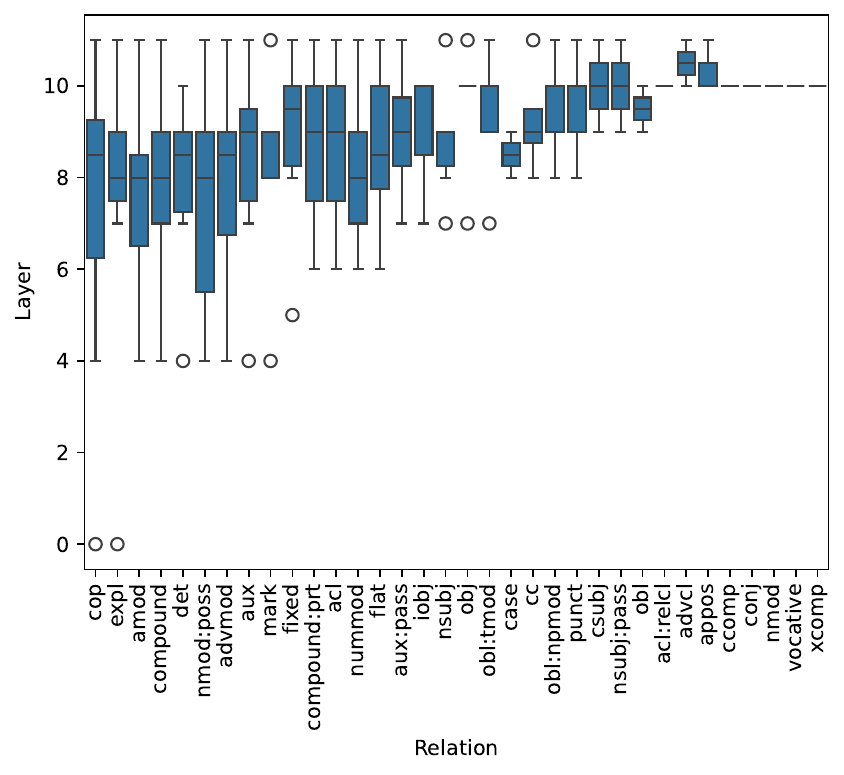}
    \caption{Distribution of location of the look-up heads we identified per UD relation across the layers.}
    \label{fig:dist-lookup-per-layer}
\end{figure}

\begin{figure}[t]
    \centering
    \includegraphics[width=\linewidth]{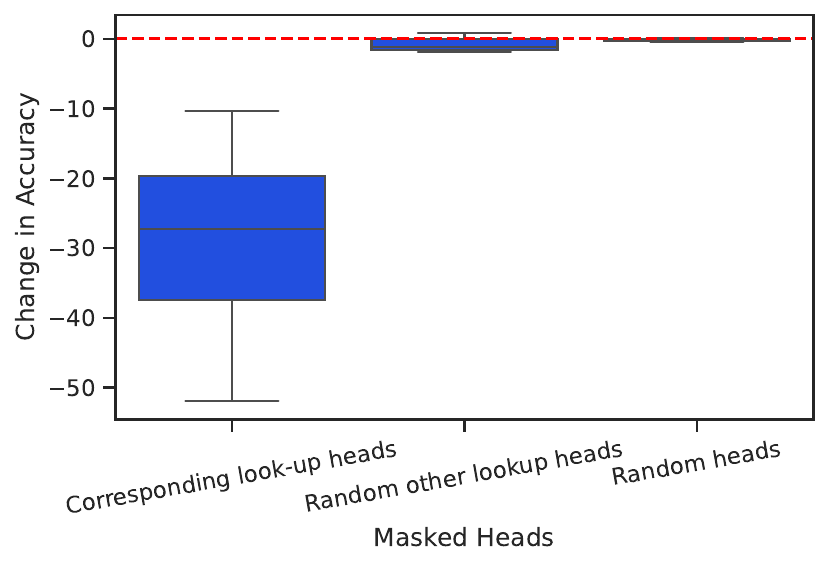}
    \caption{Effect of masking look-up heads of models fine-tuned on synthetic tasks. The boxplot shows the distribution for 10 synthetic downstream tasks.}
    \label{fig:masking-simple-finetune}
\end{figure}

\end{document}